\documentclass[conference]{IEEEtran}
\IEEEoverridecommandlockouts
\usepackage{cite}
\usepackage{amsmath,amssymb,amsfonts}
\usepackage{algorithmic}
\usepackage{graphicx}
\usepackage[warn] {textcomp}
\usepackage[edges]{forest}
\usepackage{xcolor}
\usepackage{tabularx}
\usepackage[acronym,toc,shortcuts]{glossaries}
\def\BibTeX{{\rm B\kern-.05em{\sc i\kern-.025em b}\kern-.08em
    T\kern-.1667em\lower.7ex\hbox{E}\kern-.125emX}}
\usepackage{url}

\newacronym{se}{SE}{Superintelligent Entity}
\newacronym{re}{RE}{Requirements Engineering}
\newacronym{ai}{AI}{Artificial Intelligence}
\newacronym{sai}{SAI}{Superintelligent Artificial Intelligence}
\newacronym{hlmi}{HLMI}{Human-Level Machine Intelligence}

\usepackage{turnstile}
\usepackage{adjustbox}

\renewcommand{\makehor}[4]
  {\ifthenelse{\equal{#1}{n}}{\hspace{#3}}{}
   \ifthenelse{\equal{#1}{s}}{\rule[-0.5#2]{#3}{#2}}{}
   \ifthenelse{\equal{#1}{d}}{\setlength{\lengthvar}{#2}
     \addtolength{\lengthvar}{0.5#4}
     \rule[-\lengthvar]{#3}{#2}
     \hspace{-#3}
     \rule[0.5#4]{#3}{#2}}{}
   \ifthenelse{\equal{#1}{t}}{\setlength{\lengthvar}{1.5#2}
     \addtolength{\lengthvar}{#4}
     \rule[-\lengthvar]{#3}{#2}
     \hspace{-#3}
     \rule[-0.5#2]{#3}{#2}
     \hspace{-#3}
     \setlength{\lengthvar}{0.5#2}
     \addtolength{\lengthvar}{#4}
     \rule[\lengthvar]{#3}{#2}}{}
   \ifthenelse{\equal{#1}{w}}{% New wavy $\sim$ definition
     \setbox0=\hbox{$\sim$}%
     \raisebox{-.6ex}{\hspace*{-.05ex}\adjustbox{width=#3,height=\height}{\clipbox{0.75 0 0 0}{\usebox0}}}}{}
  }

\begin{document}

%\sloppy

\title{ Superintelligence Safety:\\
A Requirements Engineering Perspective
% Superintelligence Safety - A Requirements Engineering Approach \\
% {}
}

\author{
\IEEEauthorblockN{Hermann Kaindl and Jonas Ferdigg}
\IEEEauthorblockA{Institute of Computer Technology\\
TU Wien\\
Vienna, Austria\\
\{hermann.kaindl, e1226597\}@tuwien.ac.at}
}

%\author{
%\IEEEauthorblockN{Anonymous}
%\IEEEauthorblockA{Anonymous}
%% \IEEEauthorblockN{Jonas Ferdigg BSc.}
%% \IEEEauthorblockA{Institute of Computer Technology \\
%% TU Wien, Vienna, Austria\\
%% e1226597@student.tuwien.ac.at}
%}

\maketitle

\begin{abstract}
Under the headline ``AI safety'', a wide-reaching issue is being discussed, whether in the future some ``superhuman artificial intelligence'' / ``superintelligence'' could
% even 
% harm mankind
could pose a threat to humanity. In addition, the late Steven Hawking warned that the rise of robots may be disastrous for mankind. A major concern is that even benevolent superhuman {\em artificial intelligence} (AI) may become seriously harmful if its given goals are not exactly aligned with ours, or if we cannot specify precisely its objective function. Metaphorically, this is compared to king Midas in Greek mythology, who expressed the wish that everything he touched should turn to gold, but obviously this wish was not specified precisely enough.
In our view, this sounds like requirements problems and the challenge of their precise formulation. (To our best knowledge, this has not been pointed out yet.) As usual in requirements engineering (RE), ambiguity or incompleteness may cause problems. In addition, the overall issue calls for a major RE endeavor, figuring out the wishes and the needs with regard to a superintelligence, which will in our opinion most likely be a very complex software-intensive system based on AI. This may even entail 
theoretically
defining an extended {\em requirements problem}.
\end{abstract}


\section{Introduction}

The idea of technology becoming sentient has been a common theme in the  literature and movies for decades. One might think of the famous movie ``2001: A Space Odyssey" by Stanley Kubrick, the Matrix, or the Terminator movies. These stories seem to be mostly consistent in the impression that a suddenly arising uncontrolled \gls{ai} will not mean us well, but is more likely to stand in direct opposition to humanity's goals. Furthermore, the \gls{ai}s described in books and movies are never dumb machines, but rather potent entities with cognitive superpowers far beyond the capacities of human general intelligence --- they are \textit{superintelligent}. In his book ``Superintelligence" \cite{superintelligence}, Oxford philosophy professor Nick Bostrom provides good reasons to believe that we can create such an entity.
%, and he is not the only one.

In the AI research priorities document published in
% [Russell et al. 16]
\cite{Russell_16}, apart from the desirability of safety, e.g., of self-driving cars, ``AI safety'' is addressed with regard to ``superhuman AI''. This is partly based on forecasting work on intelligence explosion and ``superintelligence'' \cite{superintelligence}, where the importance of given goals to be exactly aligned with those of mankind is emphasized. Even that may be insufficient, however, the system must also somehow be deliberately constructed to pursue them \cite{superintelligence}. 

In terms of requests like ``Make paperclips'', such a system is envisaged to take everything it can (with high ``intelligence'') and to make as many paperclips as it can make, whatever it may cost. Much like the example of king Midas
or the core theme of the movie with the title ``Bedazzled'', we view this as a requirement that is not specified precisely enough. While other references regarding ``AI safety'' can be found at \url{https://vkrakovna.wordpress.com/ai-safety-resources/}, it is mentioned nowhere that it is actually a problem with {\em requirements}.

%CR For ``AI safety'', {\em benevolence-based trust} will have to be taken into account, since its stability is contingent upon whether the trustee's actions match the goals and motivations of the truster [Lee and See 04]. The trustee is here some ``superhuman AI'', a super-intelligence based on AI-technology. 

Since mankind may eventually create a ``superintelligence'', it should rather sooner than later specify the requirements on such a system, which are traditionally the required functions, qualities and constraints. {\em Goal-oriented RE} (GORE) focusses on goals, but these are usually goals of the stakeholders rather than those of an AI system
(for GORE see, e.g., 
% [Dardenne et al. 93], [Mylopoulos et al. 99] 
\cite{Dardenne_et_al_93,Mylopoulos:1999:OGR:291469.293165}
and successor work, and the very recent mapping study \cite{Horkoff2019}). 
Hence, we propose to extend GORE theoretically in this direction for ``AI safety'', where the goals of the ``superintelligence'' based on AI technology are supposedly of critical importance.
Note, that such goals will most likely be conditions of something to achieve, in contrast to certain problem-solving algorithms where goals are defined states, see, e.g., \cite{Kaindl_et_al_95,Kaindl_and_Kainz_97}. For unidirectional search, conditions as goals can be used, in contrast to bidirectional search.

Reasoning of a superintelligence on its own goals may involve a kind of 
{\em self-awareness} \cite{jantsch:2014a}, or at least {\em self-representation} \cite{TSMC_6301747}.
It is not clear, however, what the implications are on the problem at hand.

Having a {\em glass-box view} on a system implementing a superintelligence may be helpful for keeping control of it, such as following an approach to
(object-oriented) software engineering for AI systems \cite{Kaindl_94}, \cite{Kaindl_2005}.
Also making architectural decisions upfront, e.g., for a
generic architecture \cite{ICSR_10.1007/978-3-540-44995-9_10},
may be useful in this regard.
Another approach is to model safety {\em frameworks}, see, e.g., \cite{Everitt_et_al_19}.
However, implementing a superintelligence may well involve different approaches such as deep neural networks, which are considered opaque, or approaches yet to be developed.
Hence, we take a {\em black-box view} here and propose to focus on requirements.

Strictly speaking,
the notion ``AI safety'' is misleading in our opinion, since it may also include the safety of certain AI-based systems, e.g., self-driving cars.
Hence, we prefer the notion of ``superintelligence safety''.
% in this paper.

Safety of software-intensive systems is certainly a related area \cite{ smith:11,EDCC_7780359}, but it is primarily concerned with hazards in the environment of the system and related safety risks. {\em Functional safety} as dealt with, e.g., in the current automotive standard ISO 26262 focuses on failures of functions and how to manage them.
Hence, we do not see how previous safety research could help much with regard to superintelligence safety.

The remainder of this paper is organized in the following manner. 
First, we provide some background material on superintelligence safety, in order to make this paper self-contained.
Then we view superintelligence safety from the perspective of RE, and envisage formulating requirements accordingly.
In order to address the essence of the given issue, we argue for doing RE on the superintelligence in the first place.
Finally, we motivate how the theory of the {\em requirements problem} should be extended to cover goals of a superintelligence.

\section{Background on Superintelligence Safety}

Surveys revealed that 
AI
experts attribute a moderate
% probability 
likelihood
of superintelligence emerging within a 100-year horizon, as can be seen in
% Tables~\ref{table: human} 
Tables~I
and \ref{table:superintelligence}.\footnote{PT-AI: Participants of the conference \textit{Philosophy and Theory of AI}, 2011. AGI: Participants of the conference \textit{Artificial General Intelligence} and \textit{Impacts and Risks of Artificial General Intelligence}, 2012. EETN: Sampled members of the Greek Association for Artificial Intelligence, 2013. TOP100: The 100 top authors in artificial intelligence, measured by citation index, May 2013}
Several 
AI experts where asked to estimate a year by which the 
% probability 
likelihood of having created 
% ``human-level machine intelligence''
\gls{hlmi} and superintelligence,
respectively,
reaches a certain percentage.

\def\arraystretch{1.25}
\begin{table}[t]
	\centering
	\begin{tabular}{|l|c|c|c|}
		\hline
		\textbf{} & \textbf{10\%} & \textbf{50\%} & \textbf{90\%}\\ \hline
		PT-AI & 2023 & 2048 & 2080 \\ \hline
		AGI & 2022 & 2040 & 2065 \\ \hline
		EETN & 2020 & 2050 & 2093 \\ \hline
		TOP100 & 2024 & 2050 & 2070 \\ \hline
		Combined & 2022 & 2040 & 2075 \\ \hline
	\end{tabular}
	\vspace{2mm}
%	\label{table:human_intelligence_time}
	\label{table:human}
	\caption{When will human-level machine intelligence be attained? Source: \cite[p.~19]{superintelligence}}
\end{table}

% While Bostrom points out that there are multiple ways of creating superintelligence, the idea of a self-modifying seed \gls{ai} is a promising path on which we will be focusing in this paper and since it is a software problem, software engineering methods apply in this context. 
 
Assuming the creation of a superintelligence is possible, how can we control it and avoid the doomsday scenarios so eloquently depicted by authors and film directors?
While an emerging superintelligence might not be inherently malevolent, it still poses a great risk. Without any information about the internal workings of such an entity, we have to assume that its way of `thinking' might be very different from the way a human thinks. A superintelligence most likely will not share human morals or have a concept of value at all, if the developers did not intentionally design it to have one. 

While the most terrifying scenario is a malevolent superintelligence with the intent to eradicate humanity, 
%a more likely 
another potential 
scenario is a superintelligence with no inherent motivations at all, simply following the instructions of it's creator to the best of its abilities.
% In his book, 
Bostrom describes some of the abilities a superintelligence could have under the term ``Cognitive Superpowers" \cite[p.~91]{superintelligence}. A superintelligence could excel at
% activities like 
strategic planning, forecasting, analysis for optimizing chances of achieving distant goals, social and psychological modeling and manipulation, rhetoric persuasion, hacking into computer systems, design and modeling of advanced technologies, and the list goes on \cite[p.~94]{superintelligence}. 

One can see that even equipped with only a subset of these skills, a superintelligence pursuing a goal \textit{to the best of its abilities} and without being constrained by concepts like morality or value, can cause substantial damage to its environment. Bostrom depicts this with his famous thought experiment where a superintelligence is given the simple task to make as many paper clips as possible. While the superintelligence might start off by acquiring monetary resources by predicting the stock market and building paper clip factories, it may not just stop there. Since its goal is the unconstrained maximization of the production of paperclips, it will soon discover that humans pose a hindrance to its endeavor, as we might try to stop the superintelligence from producing more paperclips or even try to shut it off. Since the superintelligence is multiple magnitudes smarter than humanity combined, humanity fails to stop the superintelligence and ultimately the whole world (the whole observable universe, in fact) will be made into paperclips.

Even if this was just a contrived example, one can see the risks of giving a superintelligence an unconstrained optimization problem.
This argument is also extended in \cite{superintelligence} to constrained optimization problems, but the details are not necessary for our paper.

% Further inquiry reveals that the same problem holds for constrained optimization problems: Even if we would instruct the superintelligence to produce exactly X paperclips, it would count them time and time again because it can never be 100\% sure that it produced exactly X paperclips. It would than gather resources to improve its reasoning capacities, leading to a similar outcome as with the unconstrained optimization problem: all the available resources on the planet (and subsequently the whole observable universe) are instrumentalized to maximize the probability of having produced exactly X paperclips \cite{superintelligence}.

\begin{table}[t]
	\centering
	\begin{tabular}{|l|c|c|}
		\hline
		\textbf{} & \textbf{2 years after HLMI} & \textbf{30 years after HLMI} \\ \hline
		TOP100 & 5\% & 50\% \\ \hline
		Combined & 10\% & 75\% \\ \hline
	\end{tabular}
	\vspace{2mm}
%	\label{table:superintelligence_time}
	\label{table:superintelligence}
	\caption{How long from human level to superintelligence? Source: \cite[p.~20]{superintelligence}}
\end{table}
\def\arraystretch{1.00}

\section{Specifying and Communicating Requirements to a Superintelligence}

Under the 
(usually unrealistic)
assumption that we knew the requirements already, `just` telling the superintelligence about them properly is the problem here.
It can be decomposed into specifying and communicating.

According to the Standard 
ISO/IEC 10746-2:2009 Information technology -- Open Distributed Processing -- Reference Model: Foundations, 7.4,
a {\em specification} is a
``concrete representation of a model in some notation''. 
In order to better understand this sub-problem of specifying requirements, let us follow
the observation in
% Kaindl and Svetinovic
\cite{Kaindl2010}
that
requirements {\em representations} are often confused with requirements {\em per se}. This confusion is also widespread in practice, as exemplified 
% this very problem is easy to see even 
in the very recent Standard ISO SE Vocabulary 24765:2017. In fact, it defines a requirement both as a ``statement that translates or expresses a need and its associated constraints and conditions'' and ``a condition or capability that must be present in a product \dots'' (dots inserted).

As it is well-known in RE, specifying requirements can be done informally, say, using some natural language, or formally, e.g., using some formal logic as the notation. Possibilities for semi-formal representations within this spectrum are many-fold.

In order to avoid {\em ambiguity} in the course of specifying requirements for a superintelligence, {\em formal} representation of the requirements should be the choice, of course, possibly in some formal logic.
{\em Grounding the logic} used in the domain will, however, still leave loopholes for the superintelligence to `misunderstand' the given specification of our requirements.
For example, for some predicate on paperclips, a grounding of what paperclips are in the real world will be necessary.

In addition, {\em incompleteness} of the specification remains an open issue. Unfortunately, no general solution appears to be feasible at the current state of the art in RE.

{\em Communicating} a given requirements specification to a superintelligence may be investigated according to \cite{4685655}, based on {\em speech-act theory} \cite{searle:69}, under 
the premise that stakeholders communicate information
% to the software engineer 
in the course of requirements engineering. 
This view was not taken for the sake of really communicating requirements
in \cite{4685655}, but for using the semantics of various speech acts to theoretically (re-)define the {\em requirements problem} originally defined by Zave and Jackson \cite{Zave:1997:FDC:237432.237434}
(see also below).
Everything is assumed to be communicated by speech acts here. However, we can only communicate {\em representations} of requirements, and not requirements {\em per se}. It is clear that the text given in the examples in \cite{4685655} {\em represents} something by describing it, the very confusion addressed in
% in our own previous work 
\cite{Kaindl2010}, where specific examples of this confusion are given.

Anyway, for communicating specifications of requirements to a superintelligence, speech-act theory could be helpful for annotating the specific kind of speech act, e.g., {\em Question} or {\em Request}.
For example, the text ``Can you produce paperclips?'' may be interpreted either way, unless specified more precisely.

Still,
we have to face that `perfectly' specifying and communicating requirements to a superintelligence may not be possible.
Even if we could, this would not help in case of malevolent superintelligence that would not necessarily satisfy requirements as specified and communicated.
Hence, let us consider building a superintelligence in such a way that these problems can be mitigated.

\section{RE on a Superintelligence}

%	According to such a new theoretical formulation of a requirements problem, we propose to actually do RE for ``AI safety''. 

In our view, building a superintelligence should certainly include RE (much as building any non-trivial system).
For specifically addressing superintelligence {\em safety}, we may ignore defining {\em functional requirements} on its superintelligent abilities.
Still, some functionality may have to be defined for operationalizing certain ``non-functional'' safety requirements, which will in the first place be cast as {\em constraints} on the superintelligence.

Working on these requirements could be done as usual according to best practice of RE.
This will involve
% interviews with 
major stakeholders (in particular, AI researchers), requirements elicitation may be done including questionnaires in the Web, etc.
While the core questions will be about requirements on superintelligence safety, of course, a bigger question should be asked in our opinion, 
what the goals and requirements are that mankind has regarding AI 
in the first place.
\begin{figure}[t]

\centering
\begin{footnotesize}

\begin{forest}
	for tree={
		align=center,
		font=\sffamily,
		edge+={thick},
		l sep'+=10pt,
		fork sep'=10pt,
	},
	forked edges,
	if level=0{
		inner xsep=0pt,
		tikz={\draw [thick] (.children first) -- (.children last);}
	}{},
	[Capability Control
		[Boxing methods]
		[Incentive methods]
		[Stunting]
		[Tripwires]
	]
\end{forest}

\end{footnotesize}

\caption{Overview of capability control techniques}

\label{fig:capability_control}

\end{figure}
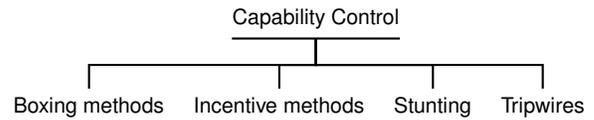

\begin{figure*}[t]

\centering

\begin{footnotesize}

\begin{forest}
	for tree={
		align=center,
		font=\sffamily,
		edge+={thick},
		l sep'+=10pt,
		fork sep'=10pt,
	},
	forked edges,
	if level=0{
		inner xsep=0pt,
		tikz={\draw [thick] (.children first) -- (.children last);}
	}{},
	[Motivation Selection
	[Direct specification]
	[Domesticity]
	[Indirect normativity]
	[Augmentation]
	]
\end{forest}

\end{footnotesize}

\caption{Overview of motivation selection techniques}

\label{fig:motivation_selection}

\end{figure*}
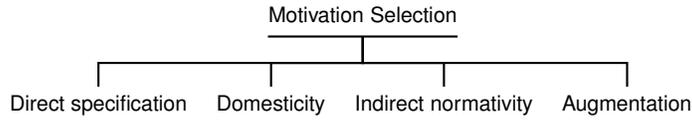

We expect that RE for {\em superintelligence safety} will pose even greater challenges than the usual practice of RE. 
After all, it is not just about conceiving such a superintelligence but about making sure that its creation will not raise uncontrollable safety risks.

Let our RE endeavor be also informed by some preliminary thought by the author who raised the issue of superintelligence safety.
In \cite[Chapter 9]{superintelligence}, {\em controlling} a superintelligence is discussed, in order to deal with this specific safety problem.
Two broad classes of potential methods are distinguished --- {\em capability control} and {\em motivation selection}.
Within each of them, several specific techniques are examined.

We review the key ideas here, starting with {\em capability control}, where Figure~\ref{fig:capability_control} shows an overview of the related techniques summarized as follows \cite[p.~143]{superintelligence}:

\begin{itemize}

\item Boxing methods\\
``The system is confined in such a way that it can affect the external world only through some restricted pre-approved channel. Encompasses physical and informational containment methods.''

\item Incentive methods\\
``The system is placed within an environment that provides appropriate incentives. This could involve social integration into a world of similarly powerful entities. Another variation is the use of (cryptographic) reward tokens. \dots'' (dots inserted)
% ``Anthropic capture'' is also a very important possibility but one that invokes esoteric considerations.''

\item Stunting\\
``Constraints are imposed on the cognitive capabilities of the system or its ability to affect key internal processes.''

\item Tripwires\\
``Diagnostic tests are performed on the system (possibly without its knowledge) and a mechanism shuts down the system if dangerous activity is detected.''

\end{itemize}

Viewed from an RE perspective, it seems as though most of what is discussed here could be elaborated as {\em constraint requirements}.
Both current theory and practice of RE will most likely be sufficient for dealing with {\em capability control} as laid out here.

With regard to the ideas on
{\em motivation selection},
Figure~\ref{fig:motivation_selection} shows an overview of the related techniques summarized as follows \cite[p.~143]{superintelligence}:

\begin{itemize}

\item Direct specification\\
``The system is endowed with some directly specified motivation system, which might be consequentialist or involve following a set of rules.''

\item Domesticity\\
``A motivation system is designed to severely limit the scope of the agent's ambitions and activities.''

\item Indirect normativity\\
``Indirect normativity could involve rule-based or consequentialist principles, but is distinguished by its reliance on an indirect approach to specifying the rules that are to be followed or the values that are to be pursued.''

\item Augmentation\\
``One starts with a system that already has substantially human or benevolent motivations, and enhances its cognitive capacities to make it superintelligent.''

\end{itemize}

%\dots

In our opinion, these ideas will be much harder to elaborate according to the current state of the art in RE
than the ideas on capability control above.
When interpreting some of them in such a way as involving 
{\em dynamic goals of the superintelligence},
GORE comes to mind.
However, can the current GORE approaches and especially the current theoretical formulation of the RE problem really cover that?

\section{Towards a New Theory of the RE Problem}

% As indicated above,
% building a superintelligence should involve Requirements Engineering.

In order to address this question, we raise further questions:
\begin{itemize}
\item What exactly is the requirements problem in this regard?
\item Can we cast it in terms of the theoretical requirements problem?
\end{itemize}

In this regard, we investigate the state of the art on formulations of the (theoretical) requirements problem.
Based on the seminal work of Zave and Jackson \cite{Zave:1997:FDC:237432.237434},
Jureta et al.~\cite{4685655} extended the formulation of the requirements problem. 
% by [Zave and Jackson 97]. 
The essence of this new formulation is as follows (where the notation is simplified for the purposes of this paper):

{\em Given:}\\
$K$	domain assumptions\\
$G$	goals\\
$Q$	quality constraints and softgoals\\
$A$	attitudes (preferences)

{\em To be found:}\\
$P$	plans

% $K,P\turnstile{s}{w}{}{}{n}G, Q, A$

\begin{equation*} 
   K,P\turnstile{s}{w}{}{}{n}G, Q, A 
\end{equation*}

% $A\turnstile{s}{w}{}{}{n}B$

An important change in this formulation is the defeasible {\em consequence} relation for non-monotonic satisfaction instead of (monotonic) entailment, based on the insight that the latter would be unrealistic with regard to practice. We consider this a useful variation of the previous formulation in \cite{Zave:1997:FDC:237432.237434}, in particular with respect to what is given in terms of goals and softgoals. Hence, this formulation elaborates the RE problem towards GORE.
Goals in GORE research are typically {\em wishes} (desires) of some agent / stakeholder.

In addition, Jureta et al.~\cite{4685655} introduced {\em attitudes} for evaluation in terms of degree of favor or disfavor of other elements in the problem formulation. While the formulation
% by [Zave and Jackson 97] 
in \cite{Zave:1997:FDC:237432.237434} 
assumed that all given ``requirements'' are compulsory, the inclusion of attitudes leads to a more realistic formulation where some are not compulsory but optional. It also covers {\em preferences} for comparison between different components of the same type, which establishes orders between components of the same type. (In particular, for each preference order in $A$ over softgoals, there is a preference order that maintains that same ordering over quality constraints that stand in the justified approximation relationship to the given softgoals defined in \cite{4685655}.) This makes it possible to define an optimum in terms of stakeholder attitudes. 

%CR This formulation was most likely inspired by {\em AI planning} research, where {\em goals} are typically achieved by {\em plans}. In [Kaindl 2000], we modeled goals directly achievable through scenario execution by viewing {\em scenarios / use cases} as plans. More specifically, these are multi-agent plans to be executed by {\em users} (major stakeholders) together with the system-to-be-built once in operation. 

From a practical point of view, we do not think that both quality constraints and softgoals ($Q$) will normally be given {\em a priori}
(as assumed in \cite{4685655}). RE should include the derivation of quality constraints from softgoals. Such derivations are actually the key task of the softgoal approach to RE.

In contrast to the original approach in \cite{Zave:1997:FDC:237432.237434},
this theoretical formulation involves {\em goals}.
However, these are 
only goals of stakeholders in RE,
while goals of the system-to-be-built itself are not included. 

Later,
Jureta et al.~\cite{Jureta:2014:RPA:2666081.2629376} formally showed the fundamental differences between standard RE (as sketched above) and RE for {\em adaptive systems}. A system is adaptive if it can detect differences between its requirements and runtime performance and can adjust its behavior to cope with such deviations. It spans design-time and run-time; design-time, because design decisions influence the range of monitored inputs for the system, and the feedback mechanisms it will have; and run-time, because these mechanisms enable the system to react to at least some changes rather than ignore them. The solution concept, called {\em configurable specification}, amounts to a set of requirements configurations and evolution requirements for switching between configurations. Each configuration is shown to satisfy all properties required by
% [Zave and Jackson 97] and [Jureta et al. 08]
\cite{Zave:1997:FDC:237432.237434,4685655}. During adaptation, the system switches from one configuration to another and does so because awareness requirements became violated during the last period of stability. 

% Another formulation for {\em agents} \dots

The latter RE problem defined for adaptive systems fits an AI-based software-intensive system most likely better than the previous RE theory. However, it still takes only goals of stakeholders into account like \cite{4685655} but not goals of the system-to-be-built itself. For more conventional software and systems, this was not necessary, of course.

Rational agents like those whose decision-making has been formally verified by Dennis et al.~\cite{Dennis2016} have goals assigned, but statically and at design-time. Such agents can be designed, e.g., using the knowledge level software engineering methodology for agent-oriented programming Tropos \cite{Bresciani:2001:KLS:375735.376477}.

% The knowledge level software engineering methodology 
Tropos
% [Bresciani et al. 01]
% \cite{Bresciani:2001:KLS:375735.376477} 
takes goals of {\em agents} into account for building agent-oriented software systems. In the course of GORE, the goals of stakeholders are broken down into goals assigned to agents. As long as this is done correctly, the goals of the stakeholders are aligned with the goals of the agents. 

However, an AI-based system may also create new goals itself at run-time 
% [Cox 07]
\cite{Cox_07}.
% Hence, 
We assume that a superintelligence will (have to) be able to formulate goals on its own. This is not covered theoretically yet in terms of 
theoretical formulations of
requirements problems.

Hence,
for treating superintelligence safety as a requirements problem, 
we propose to develop a new theoretical formulation applicable to it,
% will be developed first, 
since it involves goals of both stakeholders {\em and} of the AI-based system-to-be-built. In particular, it involves that these goals will be aligned,
and that the system-to-be-built can change them itself at run-time.
% Goals of a superintelligence, however, will most likely change in the course of its run-time. If all the goals were already known at its design-time, this could be handled using RE for adaptive systems
% [Jureta et al. 14]
%\cite{Jureta:2014:RPA:2666081.2629376}
%jointly with Tropos
% \cite{Bresciani:2001:KLS:375735.376477}. 
% However, a superintelligence will most likely also create new goals itself at run-time,
% as studied already long time ago
% by [Cox 07]
%\cite{Cox_07}. 
It will be challenging to capture all this in new theoretical formulations of extended requirements problems covering superintelligence safety.

\section{Conclusion}

In this paper, we address the potentially very important issue of ``AI safety'', in the sense of {\em superintelligence safety} \cite{superintelligence}, from an RE perspective. 
To our best knowledge, this is the first RE approach to this issue, although it may seem obvious that RE is the very discipline of choice here. 

We actually distinguish two different approaches:
\begin{itemize}
\item Specifying and communicating requirements for specific problems to a concrete superintelligence, and
\item Doing RE in the course of building a superintelligence in the sense of an AI-based software-intensive system.
\end{itemize}

Based on common wisdom of RE at the current state of the art, we tentatively conclude that `perfectly' specifying and communicating requirements to a superintelligence may not be possible.
And even if this became possible in the future, it would not help in case of malevolent superintelligence that would not necessarily satisfy requirements as specified and communicated.

Hence, we envision doing RE on a superintelligence to be built.
Such an RE endeavor could be informed by some preliminary thought on {\em controlling} a superintelligence in \cite{superintelligence}.
The first approach
% for controlling 
through {\em capability control} may be dealt with properly through {\em constraint requirements}.
The second approach for controlling through {\em motivation selection}, however, appears to go beyond the current theory of RE.
In particular, we raise the challenge of extending GORE with 
{\em dynamic goals of a superintelligence}.

After having revisited the existing formulations of the (theoretical) {\em requirements problem}, we tentatively conclude that new and extended formulations will be needed for theoretically founded RE on superintelligence.
Defining a new requirements problem for a dynamic goal model of the system-to-be-built is a challenge, but the impact on an improved understanding of problems like ``AI safety'' (in the sense of superintelligence safety) would be great.

%CR Unfortunately, applying RE theory for concrete problems faces many challenges [Kaindl et al. 2002]. 

%%%

%\clearpage

%\input{draft}

%\input{requirements}

\end{document}